# AphaVoice: Text-to-Speech Model for Aphasia Patient Simulation

Al Varrel Putra Kusuma[1] Fernando Bello[1] and Joshua Brown ✉[1]
[1] Imperial College London, United Kingdom
✉ Corresponding Author: joshua.brown@imperial.ac.uk

## ABSTRACT

While language therapy is the primary treatment for post-stroke aphasia, limited clinical resources often restrict patient access to proper care. Addressing this gap requires reliable training simulations for medical professionals. However, current AI efforts focus almost exclusively on speech processing, leaving speech generation largely unexplored. This study introduces AphaVoice, a novel text-to-speech (TTS) framework explicitly engineered to synthesize highly realistic aphasic speech. Evaluated against state-of-the-art TTS systems, AphaVoice demonstrates a superior ability to capture the clinical feature of dysfluency, outperforming existing models in temporal metrics while consistently maintaining speaker identity. By leveraging latent space shifting, the model also supports multi-speaker generation without sacrificing phonetic realism. Ultimately, AphaVoice provides a scalable, high-fidelity tool for medical training and clinical simulation related to aphasia.

## I. INTRODUCTION

Aphasia is an acquired language impairment caused by focal brain damage, resulting in difficulties in speaking, reading, and/or writing coherent sentences [1]. This condition is most commonly due to brain damage from stroke, especially ischemic stroke (IS) in the dominant left middle cerebral artery [2]. Currently, 30% of post-stroke patients suffer from aphasia [3], which poses a huge challenge for rehabilitation centers in providing suitable and timely therapy.

Speech and language therapy remains the mainstay treatment for chronic aphasia, with evidence-based guidelines recommending ten hours per week over a three-week period [4]. However, limited resources restrict patients to only about seven sessions in three months [5], which is insufficient. Training therapists is the proposed solution, but it is hindered by problems in medical education. When evaluated on a 100-point scale, over 90% of students rated their own confidence in communicating with aphasic patients at below 55, with a median confidence score of just 29 [6]. Although practice is effective, it is often hampered by practical and logistical barriers such as the communication burden placed on the patient when interacting with inexperienced trainees who lack the preparation [7]. This work addresses these problems by introducing a tool that allows students to build clinical confidence without placing additional strain on patients.

Extensive research has been conducted on aphasic speech, but the majority has focused on speech processing rather than speech generation [8, 9]. While developing automatic speech recognition (ASR) to help understand PWA is crucial, training therapists is just as important as the end goal is to help patients recover their speech abilities. A good TTS tool, working with specialized LLM would enable students to practice their communication skills with PWA anytime, anywhere, building the confidence needed for real clinical practice.

Using off-the-shelf TTS is one way to quickly build a patient simulation, with Parler-TTS [10] as a potential solution due to its ability to generate customized voices based on a prompt. Another option is XTTS-v2 [11], which can generate zero-shot speech using short examples. However, these models have specific issues: Parler-TTS struggles with generation stability [12], and models relying on an autoregressive architecture, such as XTTS-v2, often suffer from performance degradation in longer speech, which manifests as word skipping and repetition due to exposure bias [13]. The contributions of this study are as follows:

- Introduce AphaVoice, a free, open-source model for text-to-speech generation tailored for aphasia.
- Benchmark multiple text-to-speech models using aphasia-specific metrics.
- Propose latent shift as a method to effectively adding diversity and solve data scarcity problems.
- Evaluate the results of the vector-shifted AphaVoice on metrics related to emotion and accent.

## II. APHAVOICE ARCHITECTURE

### A. Fine-tuned Matcha-TTS

AphaVoice was built using Matcha-TTS [14] as the baseline model. This decision was based on three key considerations: 1) It is non-autoregressive, which prevents hallucination in speaker identity and ensures it always matches the input text; 2) It has a built-in duration predictor, which is important as aphasic speech tends to be slower; and 3) It is very lightweight with only 20.9M parameters (in comparison to 482M for XTTS [11]), making it suitable for low-resource settings common in healthcare. The fine-tuning strategy was heavily inspired by the approach utilized in EmojiVoice [15], modified for the unique acoustic requirements of this work.

Moreover, the model was fine-tuned to interpret pause (,), silent (.), and long silent (…) symbols, allowing it to simulate realistic aphasic pauses. It also includes added stuttering on multiple vowels and consonants, realistically imitating false starts and retracing. The male-speaker model was also fine-tuned using data from a patient with dysarthria (motor speech disorder that weakens the muscles used for speaking [16]) resulting in less intelligible speech for better realism. Lastly, it is equipped with the HiFi-GAN [17] V1 universal vocoder to synthesize the generated mel-spectrogram.

### B. Latent Shift and Text Manipulation

Aphasia datasets are scarce, with AphasiaBank [18] being the largest online speech database. While it contains massive multilingual data, it was not designed for TTS modelling. All its audio is spontaneous, featuring issues like background noise and overlapping voices. To address this, a solution is needed to efficiently increase data diversity without the overhead of manual data cleansing and preprocessing, as patient simulations require diverse speakers and situations for optimal practice. To solve this problem, a technique called latent shift was proposed and built on top of AphaVoice.

Several studies have utilized vector steering or latent shift to manipulate synthesized speech. This is done by extracting a vector from a fine-tuned model and using it to shift the target model. The method has been proven effective for shifting emotion [19] and accent [20]. AphaVoice supports this function, enabling latent shifts across four different emotions and six accents.

Lastly, as described previously, the model utilized tokenized pauses and stuttering. To ensure all generated speech matches the PWA profile, it includes a built-in text manipulation layer that pre-processes the input text to fit the AphaVoice model. It was designed with a severity parameter (ranging from 0 to 1) that influences the probability of adding in-between pauses (by inserting pause symbols), filler words ("uh," "um…"), and stuttering ("p… pa, pass"). Fig 1 shows an overview of the AphaVoice architecture.

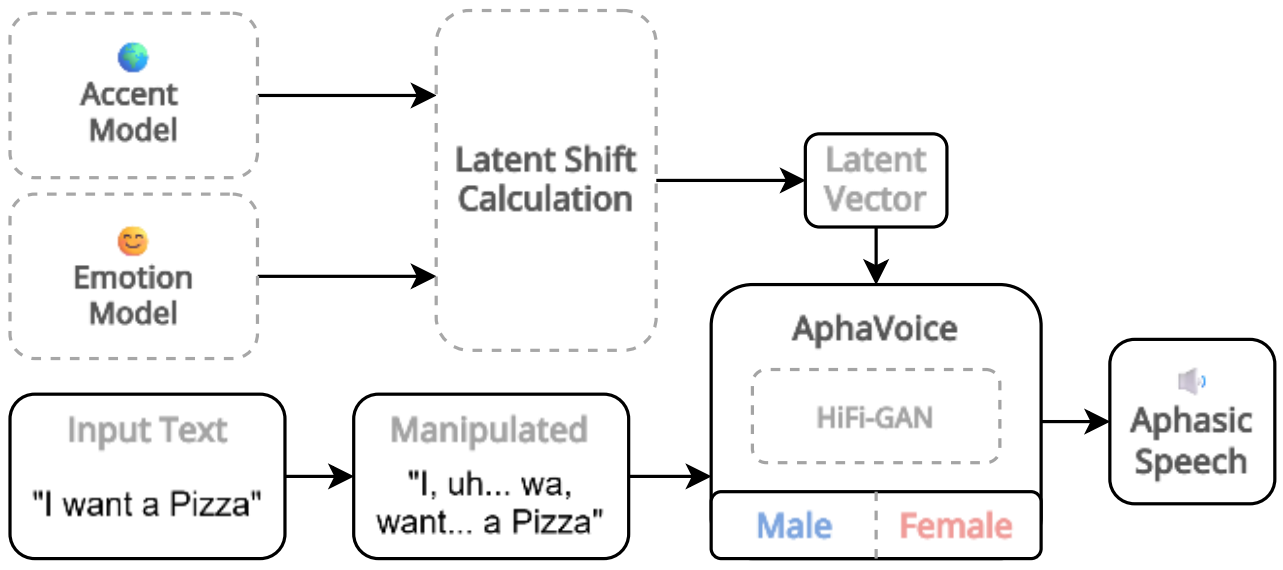

*Fig. 1. Overview of AphaVoice architecture.*

## III. Experiment Protocol

### A. Dataset Pre-Processing and Fine-Tuning

The model was fine-tuned from the multi-speaker VCTK checkpoint made available by Matcha-TTS. The final list of speakers was cut down to two to reduce leakage between speakers in inference. The training data was taken from AphasiaBank, specifically the APROCSA [21] dataset with speaker aprocsa1738a (male) and aprocsa1944a (female).

AphasiaBank encodes its transcripts as .CHA files that are accompanied by the session videos. The first pre-processing step was to crop the videos into segments, ensuring each segment matches its corresponding transcript. Each conversation was time-stamped to allow for precise cropping, after which the video was reformatted to a .wav audio file and the scripts were saved as .txt files with matching filenames. Following this, some samples were disregarded, particularly those with voice overlaps, awkward crop timing, or insufficient duration.

After that, the script was preprocessed to remove speech-specific symbols (e.g., &-, [/]). While these symbols were made specifically for aphasic speech, using them carelessly for text-to-speech fine-tuning could cause the model to break, as they were not introduced in the Matcha-TTS baseline. However, several cues (e.g., gasping) were kept and assigned to new symbols.

Next, the audio is pre-processed. First, depending on the duration of the pause, one of three symbols is added: a comma (,), a period (.), or an ellipsis (…). This process is critical for capturing the pauses made by an aphasic patient, as Matcha-TTS is not designed to handle them natively (it would read the symbols literally, vocalizing them as "dot dot dot"). Afterward, each audio file is resampled to 22050Hz, and the mel-spectrogram statistics (mean and standard deviation) are recalculated. These steps are important to fully capture the aphasic speech profile.

Finally, the data is formatted for Matcha-TTS fine-tuning and split 95:5 for training and validation. The combined two-speaker dataset is then shuffled and used for fine-tuning. AphasiaBank is a spontaneous dataset, so it naturally has a slower pace with more pauses, making it ideal for a realistic patient simulation. Also, the number of symbols increased from 198 to 205 to accommodate aphasic speech cues. The fine-tuning was performed on an RTX 6000 GPU with 24GB of RAM and took 8.5 hours for 1000 epochs.

### B. Sub-model Fine-Tuning & Vector Extraction

Alongside AphaVoice, two sub-models were built for subsequent latent shift experiments. Similar pre-processing methods were applied before fine-tuning. For emotion shifting, the Emotional Speech Dataset (ESD) [22] was used instead of the more popular IEMOCAP dataset. This was because IEMOCAP was recorded for theatrical purposes and does not fit the goal of patient simulation, as its emotions tend to be over-emphasized [23]. This dataset contains four emotional states: happy, angry, sad, and surprise. For accent shifting, L2-ARCTIC [24] was used, which contains six different accents: Vietnamese, Mandarin, Korean, Indian, Spanish, and Arabic. Both datasets are scripted. This was a deliberate choice to accurately extract the speech profiles for the AphaVoice baseline.

The vector extraction is computed as the difference between the parameter sets of the fine-tuned models (accent and emotion) and the base model (the Matcha-TTS VCTK checkpoint). The extraction formula is defined as follows:

$$\tau_{accent/emotion} = \theta_{fine-tuned} - \theta_{base} \quad (1)$$

In this formula, $\tau_{accent/emotion}$ denotes the extracted vector, which represents a direction in the model's parameter space that creates a shift in the generated voice. Here, $\theta$ represents the model's parameters. Dividing the two models results in a vector, consistent with prior studies [20]. This method was then applied to all six accents and four emotions to generate ten different vectors that form the basis for a comprehensive evaluation of the proposed method.

During inference, the extracted vector of the model with an α parameter that controls the shift's strength was added. This parameter can also be continuously controlled and tweaked to find the right balance. An alpha value that is too large can cause a catastrophic shift, producing only a buzzing sound as a result. The inference formula is as follows:

$$\varphi_{accent/emotion} = \theta_{base} + \alpha \cdot \tau_{accent/emotion} \quad (2)$$

The outcome of this inference is a shifted speech ($\varphi_{accent/emotion}$) that produces an output perceptually closer to the target accent or emotion profile. These shifted speeches were then evaluated as a method to multiply aphasic speakers.

### C. Model Benchmark

Multiple current-generation off-the-shelf models were selected for benchmarking as there is currently no TTS model made specifically for a clinical condition to the best of the authors' knowledge. Two models, XTTS-v2 and Parler-TTS were selected based on their customization capabilities. Three other models (VITS [25], Suno's Bark [26], and Microsoft's SpeechT5 [27]) were included because they are commonly used in various applications, including simulation. Finally, the base Matcha-TTS was also included as a baseline for our ablation study.

**XTTS-v2 and Parler-TTS**. XTTS-v2, developed by Coqui AI, is a zero-shot TTS model. For this model, a single audio sample from speaker aprocsa1738a was utilized to condition the model. Parler-TTS, an open-source model from Hugging Face, generates custom voices via descriptive prompts. The specific prompt used was: "A male speaker with aphasia delivers a monotonous, low-pitched, and exhausted voice. He speaks very slowly and carefully in a quiet room with a completely flat tone."

**VITS, SpeechT5, and Bark**. These three models represent widely adopted, general-purpose TTS systems. Although not engineered for clinical contexts, they establish a performance baseline against common alternatives. No specific settings were changed on these three models.

**Matcha-TTS and AphaVoice**. Matcha-TTS with VCTK checkpoint and a male speaker was selected for comparison. For AphaVoice, both male and female speakers were used to benchmark the overall performance of the model.

### *D. Evaluation Metrics*

**Clinical Metrics.** The auditory-perceptual rating of connected speech in aphasia (APROCSA) provides a comprehensive set of 27 metrics for the clinical assessment of perceptual features in aphasic speech. However, not all of these metrics are applicable to TTS benchmarking (e.g., jargon, neologism) and some are too complex to implement (e.g., off-topic speech, dysarthria). To this end, the metrics selected for TTS benchmarking are: words per minute (WPM) to cover reduced speech rate, and pause ratio, which combines the APROCSA metrics for pauses between and within utterances. Other metrics were also implemented in the model but are not included in the benchmarking because models other than AphaVoice do not support them. These are false starts, abandoned utterances, perseverations, and effortful speech.

**Intelligibility Metrics**. Two standard intelligibility metrics were also added: word error rate (WER) and character error rate (CER). Aphasia frequently co-occurs with dysarthria, which causes speech to sound slurred, choppy, or mumbled. Consequently, utterances may be truncated or abandoned. This is one of the biggest challenges in handling PWA and is a crucial profile that a TTS model needs for clinical simulation. To evaluate this, OpenAI Whisper model [28] (Base-English, faster-whisper implementation) was used to transcribe the speech generated by all eight models. The results were then compared. However, rather than aiming for the lowest score, higher scores are needed, which would prove that the fine-tuning successfully "broke" the ordinarily very fluent model. It must also be noted that the score should not be extremely high, as the generated speech must still be intelligible to the intended users: medical students.

**Speaker Consistency**. To assess speaker consistency, the speaker encoder cosine similarity (SECS) metric was used. This metric is a good fit because it explicitly ignores the content of the sentence and instead focuses on the speaker's voice characteristics, such as pitch and vocal tract. While this metric is commonly used for biometric security, the same method has been applied to evaluate speaker consistency [11, 29]. For the encoder, ECAPA-TDNN [30] was used, accessed through the SpeechBrain library [31]. It converts the raw audio files into mel-spectrograms and uses 1D convolutions to scan across the audio. The function returns a value between -1 (opposite) and 1 (perfect match).

**Sub-Model Metrics**. Two additional models were created for the latent shift experiment: emotion and accent. For emotion model, two metrics were used: pitch (F0) and RMS energy. These metrics are commonly used in classifying emotion in speech [32, 33]. For the accent model, VoxProfile [34] model was used, which is an English-accent classification model. Its performance was evaluated by calculating the average softmax posterior probability for the target accent label over 50 utterances. This evaluation methodology was adopted from prior research in the domain [20].

**Latent Shift Evaluation.** Finally, it is important to characterize the behavior of the vectors introduced during latent shifting [35]. For this, UMAP was used to visualize the distribution of ECAPA-TDNN extracted features for emotion and accent models. The calculation is defined as:

$$\Delta d = \theta_{base} - \theta_{accent/emotion} \tag{3}$$

where $\theta$ was defined as the model. This ensures the string of text for the speech does not influence the results. In addition to this, the same metrics used to evaluate the sub-models were also used to analyze the changes these vectors make to the model. However, the analysis focuses on the delta from the baseline to evaluate the latent shift implementation.

For all experiments, fifty utterances were produced using the selected models. This amount was deemed statistically useful for testing the defined metrics. Twenty-five of these scripts were generated using a local LLM, while the rest were taken from an unseen AphasiaBank dataset. Some scripts included pause punctuation and filler words (e.g., "uh…" and "um,") to test how each TTS model handles them. The exact scripts and generated audio files are available on supplementary materials and GitHub for testing and evaluation.

## IV. Results

### *A. Aphasic Speech Evaluation*

For WPM, AphaVoice scored 76.57 for the male speaker and 82.37 for the female speaker. This is more than 30% slower than the next slowest model (XTTS-v2) and 70% slower than the base Matcha-TTS model. These results align with latest study for PWA [36], as their speech rate typically sits between 61 to 88 WPM, compared to around 140 WPM for neurologically healthy individuals in the same age range (control, 54-89 years old). It is important to note that referred study used spontaneous speech data for evaluation, whereas this study TTS models generated scripted speech, which naturally result in higher scores.

Another prominent characteristic of speech in PWA is the prevalence of pauses, both within and between utterances. One study [37] shows that the mean silent proportion in aphasic speech is around 61% (non-fluent) and 50% (semantic), compared to just 21% in healthy individuals. This finding is derived from spontaneous speech, which inherently have longer pauses. AphaVoice scored the highest on this metric, with a mean score of 48.9% for the male speaker and 55.16% for the female. While Bark was the second highest with 47.6%, Fig. 2 reveals a high variance in its distribution, indicating inconsistent pausing. A similar variance is present in AphaVoice, where the female model has higher variance but also a higher mean pause ratio. Crucially, AphaVoice shows good pause-ratio that aligns with clinical observation.

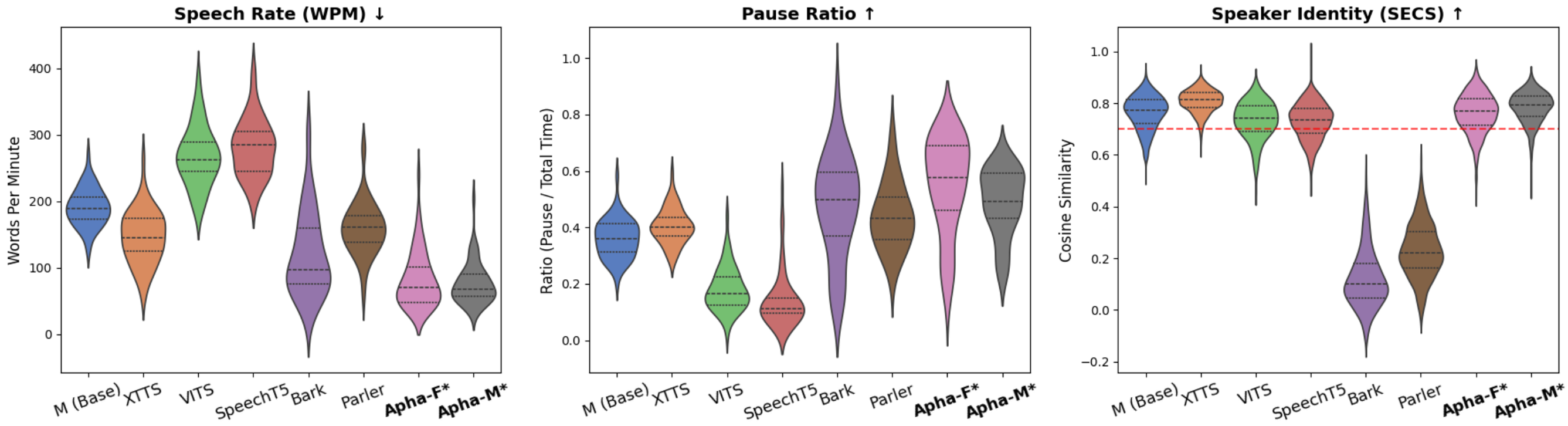

*Fig. 2.* *AphaVoice benchmark on aphasia-specific and speaker consistency metrics. Results are presented with result distribution, median, and quartile presented for better comparison. ↑ indicates that higher score is better, while ↓ indicates that lower score is better.*

Finally, since this model was built for patient simulation, it needs consistent speaker identity. This is particularly important because aphasia is highly related to a speaker's individual characteristics, making any switch in vocal patterns intolerable. For this reason, a SECS analysis was done with a goal of achieving a minimum mean score of 0.70, a threshold identified from prior studies [11, 38]. On this metric, AphaVoice scored a mean of 0.78 for the male model and 0.76 for the female model, proving consistent generation.

### *B. Intelligibility Comparison*

The primary speech profile of PWA is in producing coherent speech. This is typically shown in patients repeating words several times, having false starts (e.g., "sch-" leading to the word "school"), using fillers (e.g., "uh," "um"), and having disrupted language production (slurred or mumbled speech). To evaluate this, WER and CER were used. However, the scores should be interpreted inversely, with a higher score being better than a lower one. This is because aphasia impacts an ASR model's ability to transcribe correctly. Prior research [9] have analyzed the range of WER for aphasic speech with the result sitting in the range of 20-60% and a mean of 38%. This value will be the target range of errors for AphaVoice.

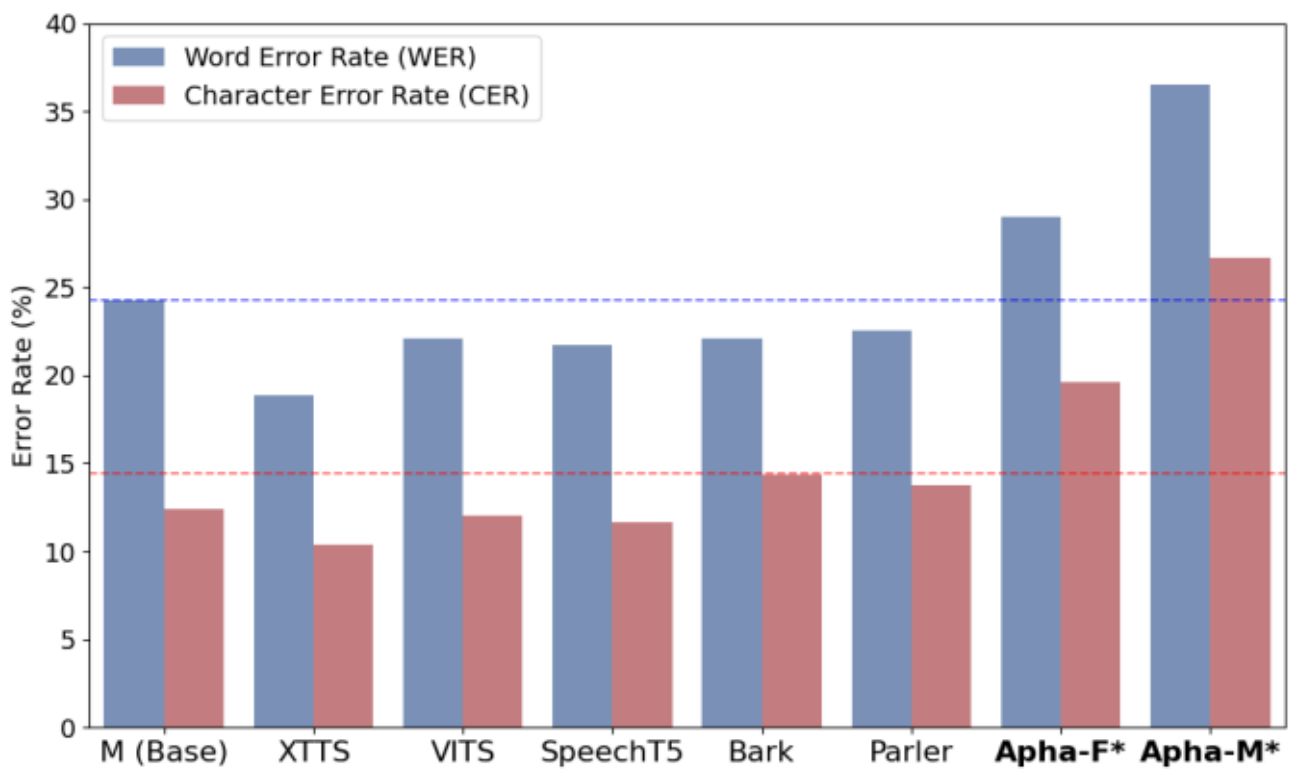

*Fig. 3. Intelligibility degradation evaluation. Results are presented as mean of combined corpus. Ticked line highlighted the second-best model after AphaVoice.*

As illustrated in Fig. 3, eight models were compared and evaluated using the same method. AphaVoice male and female attained the highest WER scores (36.5 and 29, respectively) and CER scores (26.7 and 19.6, respectively). Importantly, compared to the base pre-trained Matcha-TTS (male), AphaVoice's (male) WER and CER increased significantly by 50.21% and 115.32%, respectively (Mann-Whitney U Test: p=0.00004 for CER; p=0.00128 for WER). This indicates a successful fine-tuning, where the model now expresses an aphasic profile in its generated speech. While this is a huge increase, the speech generated remains highly intelligible and close to the target value of 38%.

To verify that the degraded transcription results reflect genuine aphasic speech rather than synthesis failure or vocoder artifacts, a further decomposition of the ASR errors was conducted. The analysis reveals that the errors overwhelmingly stem from deletions (43–49% of total errors), which correspond to omitted fillers (e.g., “uh,” “um”) and agrammatism, alongside substitutions (31–35%), which indicate successfully learned phonemic paraphasias (e.g., transcribing the intended word “departs” as “depots”). Conversely, insertions, which typically indicate synthetic acoustic noise, account for just around 20% of the total errors, maintaining a consistently low absolute rate of 5.8–7.8%.

### *C. Sub-Model Evaluation*

Two models were built alongside AphaVoice for vector extraction. For the emotion model, it focused on pitch (F0) and RMS energy. The results for both male and female models are presented in Fig. 4. As expected, the 'sad' emotion consistently demonstrated the lowest pitch and RMS energy, aligning with prior studies on low-arousal states [33]. Conversely, 'happy' exhibited the highest pitch but relatively lower RMS energy compared to high-arousal negative emotions. This reflects the acoustic profile of 'quiet happiness' or contentment, characterized by lower acoustic intensity but a dynamic pitch variance [39]. 'Angry' and 'surprise' have higher RMS energy because these emotions are spoken with much higher intensity. Finally, the pitch variance for 'anger' is lower due to the consistently sustained high pitch of the speech itself, explaining this result.

For accent, the mean classifier probability is compared between the base model and the sub-model. Since Matcha-TTS uses VCTK as its base, it has a UK English accent as its baseline (53.51% for female and 35.87% for male probability). This explains the lower results for the base model when classifying other accents, with Spanish (mapped as Romance in VoxProfile) having the highest score among them. Aside from that, the probability for all other accents is consistently under 3%, which is logical since, except for Spanish, the other accents are from Asia and the Middle East.

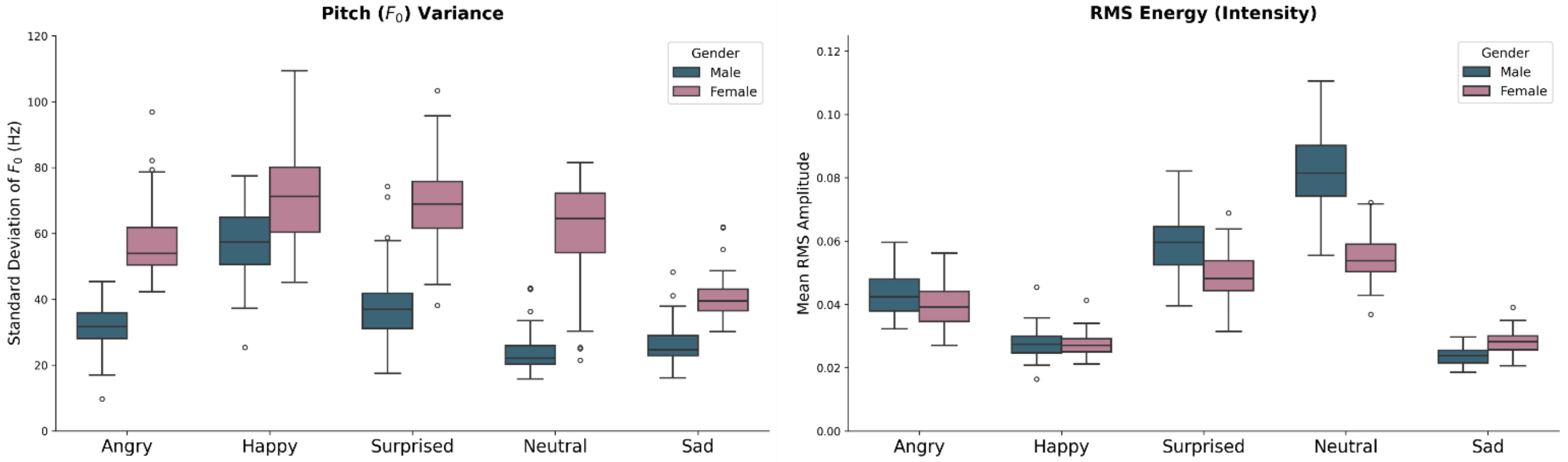

*Fig. 4. **(Left)** Pitch ($F_0$) Evaluation **(Right)** RMS Energy Evaluation. Result is presented as median with quartile (Q1 and Q3) as error bars.*

After fine-tuning, the probability increased substantially, with Indian, Korean, Arabic, and Mandarin all exceeding 89%. For Spanish, it also increased substantially, with the male model reaching 77.99% and the female 62.24%. This result is logical, as VoxProfile consolidates Romance accents from multiple countries. The same applies to Vietnamese, which is grouped under Southeast Asia (SEA) with other distinctive accents. Despite this grouping, every accent showed a clear increase in probability, demonstrating that the model captured the target accents well.

These findings were important for the latent shift experiment. As for it to be considered successful, the shifted inference must move in the same direction as the vector. For example, applying the Indian accent vector should correspondingly increase the probability for the Indian accent, while adding "happy" vector should elevate the pitch.

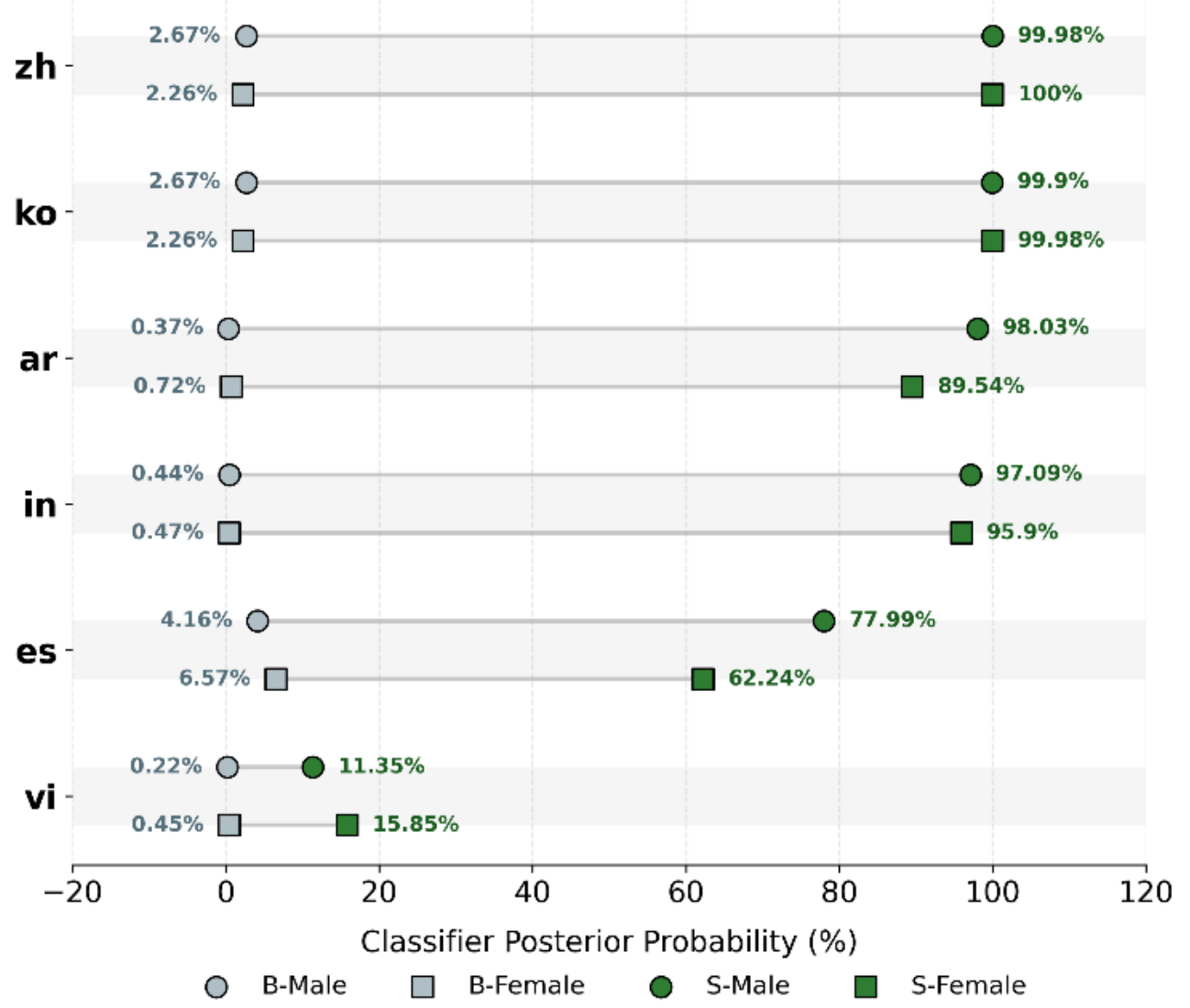

*Fig. 5. Sub-model Accent Evaluation. Base model (B) and sub-model (S) are represented by squares for female speakers and circles for male speakers. Gray = prior probability, green = final probability. For the accents: zh=Mandarin, ko=Korean, ar=Arabic, in=Indian, es=Spanish, vi=Vietnamese.*

### *D. Latent Shift Experiment*

For the final experiment, the ten extracted vectors were individually applied to AphaVoice during inference to observe the changes in its output. Prior to quantitative evaluation using the same metrics as the sub-models, it is crucial to visualize the direction of each vector's shift. This ensures that each vector behaves as intended and moves in a distinct direction without overlapping with others.

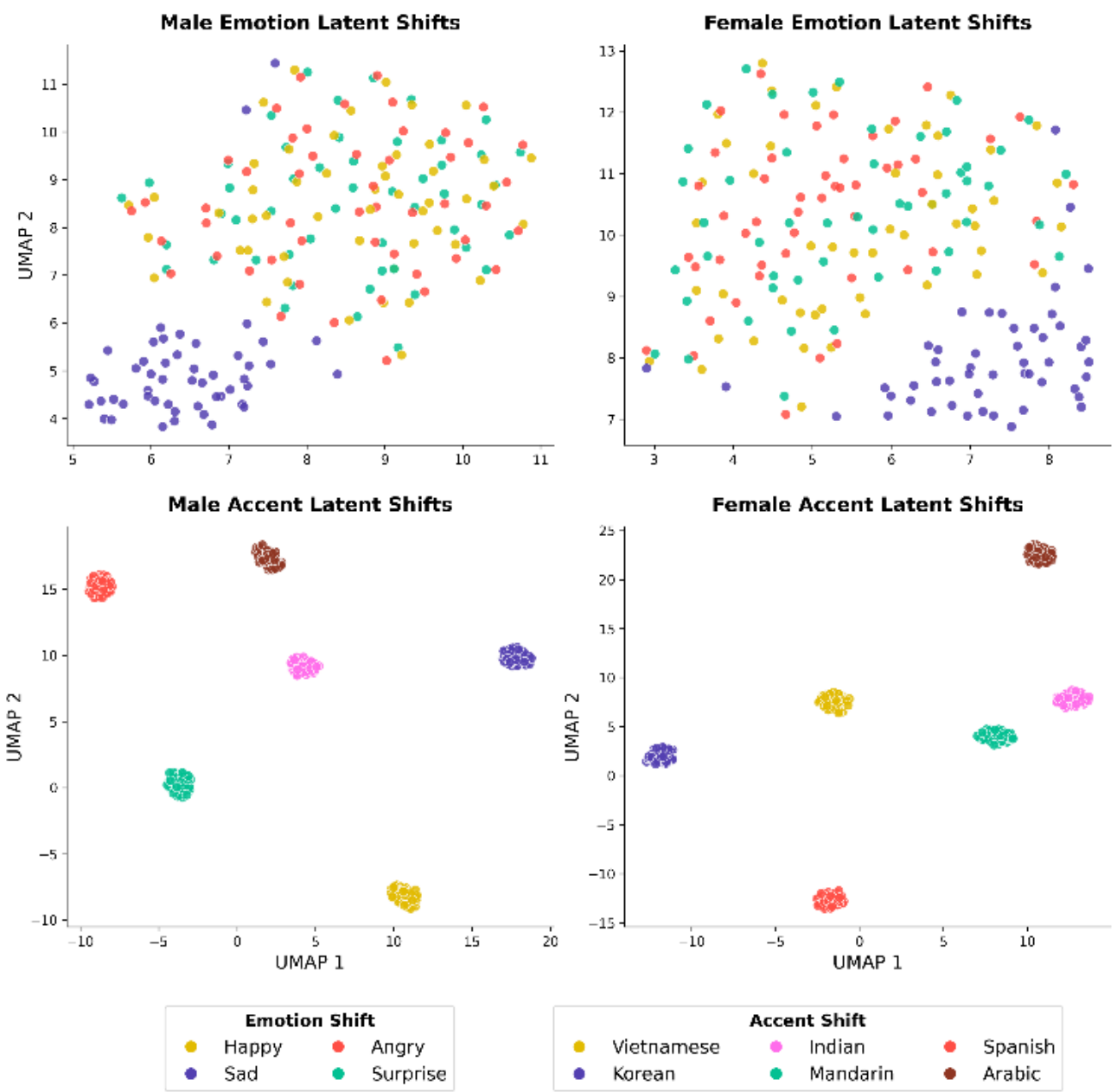

*Fig. 6. UMAP visualization for latent shift in AphaVoice. **(Top)** shows the emotion and **(Bottom)** shows the accent mapping.*

As illustrated in Fig. 6, the accents form distinct, well-defined clusters, while the emotions are jumbled and entangled. This outcome was a direct reflection of the sub-models' behavior. Accent acts as a global, systemic phonetic shift, as vowel spaces, consonant formations, and rhythmic timing are consistently altered. This resulted in clearly defined clusters, as one accent is fundamentally different from another. Conversely, emotions were not as clearly defined, especially since the three selected emotions (happy, angry, surprise) were all high arousals. This resulted in the three being jumbled and overlapping, with only 'sad' forming a clearly separated cluster.

The next step was quantitative evaluation of the shifted AphaVoice models using the same methodology applied to the sub-models. It should be noted that the primary objective of this experiment was to measure the shift rather than the base score. As shown in Table 1, all six accents for both male and female models produced positive results, evidenced by increased prediction probabilities. This confirms that the latent vectors not only behave as expected but also induce a shift in the correct direction within the model's feature space.

An additional observation was made regarding the base speakers. AphaVoice base female speaker is reported as a White woman, but she has a notable Asian accent, explaining the higher base probabilities for Korean, Mandarin, and Vietnamese. However, even though the model's base male speaker is identified as a White man with no notable Asian accent, it still shows high baseline probability. While this cannot be definitively confirmed in the absence of ground truth, it is worth noting. The classifier (VoxProfile) was trained for healthy speech, and it is plausible that the phonations of aphasic speech led to misclassifications.

The East Asian accents, especially Mandarin, were strongly boosted, with shifts of +15.33 for the male model and +16.72 for the female model. However, the impact of some vectors, such as the Indian accent, was negligible. This is interesting, as these accents were highly distinct in the sub-model but showed very different results as vectors. Qualitatively, the four generated speech samples (base, Mandarin, Indian, Korean) were all distinctly different, adding further evidence for the latent shift's effectiveness.

The other three accents yielded more limited success. The Spanish and Vietnamese vectors showed greater efficacy in the female model (both with >4% increases), whereas the Arabic vector performed better in the male model (+4.6%). Nonetheless, when listened to carefully, the difference is not as obvious as with the three previously mentioned accents. This is likely due to predominant aphasic speech profile which makes it harder to differentiate subtle accent changes.

*Table 1. Shifted AphaVoice accent evaluation. Results are presented as mean probability from 50 utterances of generated speeches.*

| Gender | Language | Accent Probability (%) | |
|---|---|---|---|
| | | Base Score | Accent Shifted |
| Female | Spanish | 15.67 | 20.33 (+4.66) |
| | Indian | 0.49 | 0.77 (+0.28) |
| | Arabic | 3.11 | 5.72 (+2.61) |
| | Korean | 56.14 | 65.83 (+9.69) |
| | Mandarin | 56.14 | 72.86 (+16.72) |
| | Vietnamese | 15.42 | 20.02 (+4.6) |
| Male | Spanish | 25.73 | 25.9 (+0.17) |
| | Indian | 0.38 | 0.48 (+0.1) |
| | Arabic | 9.37 | 13.97 (+4.6) |
| | Korean | 49.42 | 61.75 (+12.33) |
| | Mandarin | 49.42 | 64.75 (+15.33) |
| | Vietnamese | 1.8 | 3.12 (+1.32) |

For the emotion model, the initial observation is substantial discrepancy in pitch (36.11) and RMS energy (0.0265) between the male and female models. A primary reason for this is the presence of noise in the female model's output. When presented with a long pause symbol (…), the male model produces true silence, whereas the female model generates huffing and gasping sounds. These were intentionally preserved to mimic PWA pauses, but they caused a large shift in the two metrics under evaluation.

To prove that the latent shift works, the key is to observe the change in values, as they should mirror the sub-model's speech profile. For both models, 'sad' has the lowest pitch compared to the other emotions, mimicking the sub-model. The same is true for 'happy', which has the highest pitch in the male model (29.73) and is a close second in the female model. The other two emotions remain closer to neutral, which also show a similar profile to the sub-model.

The RMS energy demonstrated more mixed results than pitch. For example, 'sad' had the highest energy in both shifted models, which was the opposite of the sub-model. Conversely, the 'happy' emotion correctly mimicked the sub-model by having the lowest RMS energy. The 'surprise' emotion had the highest RMS energy in the male model (followed by 'angry'), matching the sub-model, though it was ranked behind 'sad' for the reason explained. In the female model, the 'angry' and 'surprise' values were reversed but still relatively close to neutral. Overall, the vector movements generally followed the sub-model, with the noted exceptions.

*Table 2. Shifted AphaVoice emotion evaluation. Numbers under header are base result for neutral emotion. Results are presented as mean value for both Pitch and RMS energy.*

| Gender | Emotion | Pitch ($F_0$) | RMS Energy |
|---|---|---|---|
| | | *F: 63.28*<br>*M: 27.17* | *F: 0.0684*<br>*M: 0.0949* |
| Female | Angry | 56.03 | 0.0751 |
| | Sad | 42.32 | 0.0795 |
| | Surprise | 60.57 | 0.0685 |
| | Happy | 57.49 | 0.066 |
| Male | Angry | 26.49 | 0.0971 |
| | Sad | 23.96 | 0.1057 |
| | Surprise | 24.79 | 0.1029 |
| | Happy | 29.73 | 0.0919 |

## V. Limitation & Discussion

Although a clinical mean opinion score (MOS) has not been obtained, the technical evaluation presented here demonstrates the model's performance. However, all generated audio and the model have been made publicly available, enabling further research and testing by others.

A notable omission from this study is the evaluation of speech naturalness using metrics such as UTMOS. This was a deliberate choice, as current naturalness benchmarks are predicated on fluent speech corpora like VCTK. To the best of the authors' knowledge, no comparable model trained on PWA speech provides a baseline for clinical authenticity. Using a standard fluency-based metric would unfairly penalize the model's intentional stutters and mistakes. This also presents an opportunity to create a new benchmark with aphasic speech as its foundation.

In conclusion, this study presents a novel TTS model designed for aphasia. This model holds significant potential for applications in medical education, particularly in patient simulation. To be fully deployed, it must be integrated with two other models (ASR for transcription and an LLM for script generation) to enable continuous conversation. As a clinical trial has not yet been conducted, researchers should exercise judgement when using this model.

## Acknowledgement

A.V.P.K was supported by the Indonesia Endowment Fund for Education Agency (LPDP) for his study.

## AVAILABILITY

The AphaVoice model weights, configuration, and scripts are publicly available for research and deployment on Hugging Face: https://huggingface.co/varrelkusuma/AphaVoice.

## AI-GENERATED CONTENT DISCLOSURE

The authors acknowledge the use of Google's Gemini 3.1 Pro for grammar check, paraphrasing and coding support. All generated material was reviewed, verified, and edited by the authors, who take full responsibility for the final content.